\documentclass[11pt,a4paper]{article}
\usepackage[hyperref]{acl2020}
\usepackage{times}
\usepackage{latexsym}
\usepackage{graphicx}
\usepackage{amsmath}
\usepackage{CJKutf8}
\usepackage{placeins}

\usepackage{microtype}

\aclfinalcopy 

\title{Evolution of Part-of-Speech in Classical Chinese}

\author{Bai Li \\
  University of Toronto \\ 
  Vector Institute for Artificial Intelligence \\
  \texttt{bai@cs.toronto.edu} \\}

\date{}

\begin{document}
\begin{CJK*}{UTF8}{gbsn}
\maketitle
\begin{abstract}
Classical Chinese is a language notable for its word class flexibility: the same word may often be used as a noun or a verb.
\citet{bisang-precategorical} claimed that Classical Chinese is a precategorical language, where the syntactic position of a word determines its part-of-speech category.
In this paper, we apply entropy-based metrics to evaluate these claims on historical corpora.
We further explore differences between nouns and verbs in Classical Chinese: using psycholinguistic norms, we find a positive correlation between concreteness and noun usage.
Finally, we align character embeddings from Classical and Modern Chinese, and find that verbs undergo more semantic change than nouns.
\end{abstract}

\section{Introduction}

Classical Chinese (or Literary Chinese) is the written form of the Old Chinese language from the 5th century BC until the 3rd century AD, and continued as a literary tradition until the 20th century. Compared to Modern Chinese, it is extremely compact.

Typologically, Classical Chinese is an isolating language, with almost all words consisting of one character. There are no inflectional markings on nouns or verbs, and only limited derivational morphology. It has extensive zero-derivation, so that both action-denoting and object-denoting words can be used as either a noun or a verb. \citet{bisang-precategorical} argues that Classical Chinese is precategorical: specific lexical items lack a noun/verb distinction, and instead, the distinction is determined entirely by the syntactic position.

An example of word class flexibility is the word 信 ({\em xin}), which can be used as an intransitive verb, transitive verb, or noun:

\begin{itemize}
    \item Intransitive verb “to be trustworthy”: 智士者未必{\bf信} (a wise knight is not always {\bf trustworthy})
    \item Transitive verb “to believe”: 人皆{\bf信}之 (they all {\bf believed} him)
    \item Noun “confidence”: 无道得小人之{\bf信}也 (he has no way of achieving the {\bf confidence} of the commoners)
\end{itemize}

Word class flexibility exists in many languages, including English, but is generally limited to only some lexical items. If it were indeed the case that any word can be used both as a noun and verb, then that would make Classical Chinese typologically a very unusual language. One must then wonder how such a system came about, and how it can resolve ambiguity sufficiently to serve the everyday communicative needs of its speakers.

In this paper, we evaluate whether Bisang's claims can be supported by corpus-based evidence. It turns out that while we can find many flexible words, there are also many words that appear to be inflexible, at least in our corpus. Through entropy-based measures, we find no evidence to support that syntactic position determines part-of-speech any more in Classical Chinese than in Modern Chinese (which we treat as a control).

There are, however, some differences between nouns and verbs in Classical Chinese. First, we find evidence for another claim made in Bisang's paper, that concrete words are used more as nouns and abstract words more as verbs. Second, we find that verbs undergo more semantic change from the Classical to Modern Chinese languages.

\section{Related work}

\subsection{Word class flexibility}

A fundamental question in linguistic typology is whether a language has a noun-verb distinction. Various languages such as Mundari, Riau Indonesian, and the Salish family are claimed to lack a noun-verb distinction, but other linguists question whether this is really the case \citep{flexible-ch1, haspelmath2012}. In any case, it is important to define precisely what is meant for a language to lack a noun-verb distinction.

\citet{evans-osada} give three criteria: compositionality, bidirectionality, and exhaustiveness. Compositionality means any morphological or syntactic rule must apply equally to both nouns and verbs. Bidirectionality means it is not sufficient just for nouns to be used in verbal positions, but it must also be possible for verbs to be used in noun positions. Exhaustiveness means the flexibility must be applicable to the whole lexicon, and not just a few examples. If a language satisfies all three criteria, then it can be said to lack a noun-verb distinction.

\subsection{Part-of-speech induction}

Unsupervised part-of-speech induction has been a topic of research for several decades: \citet{unsupervised-pos-induction-survey} gives a good survey. These methods attempt to assign words to different clusters that share a POS tag; however, they are usually linguistically unmotivated, and evaluation is problematic, especially for languages for which the existence of a noun-verb distinction is contentious. \citet{mistica-indonesian} uses the equivalent combinatorics criterion to cluster words in Indonesian based on which suffixes are permissible on a stem morpheme. This method is not applicable to Classical Chinese because of its highly isolating morphology.

\subsection{Classical Chinese NLP}

Various natural language processing methods have been applied to Classical Chinese. \citet{pre-qin-freq} computed basic frequency and POS statistics on a corpus of 25 pre-Qin Chinese texts, including character-level entropy. \citet{crf-pos-tagger} used conditional random fields (CRFs) for word segmentation and POS tagging, although their corpus was linguistically much closer to Modern Chinese.

Deep learning has also been applied to Classical Chinese with some success. \citet{bilsm-pos-tagger} used a BiLSTM + CRF model for sentence segmentation, word segmentation, and POS tagging on {\em Siku Quanshu}, a collection of over 10,000 books. \citet{bert-sentence-seg} trained a BERT+CRF model for sentence segmentation. In most cases, it is not possible to replicate these models because the authors do not publicly share their training data.

\section{Data and Methods}

We use several Classical Chinese corpora. First, the Kyoto corpus, part of the Universal Dependencies project, contains POS-tagged dependency trees for 74k characters of the four pre-Qin Confucian classics: Analects, Mencius, Doctrine of the Mean, and Great Learning \citep{kyoto-corpus}. Also part of the Universal Dependencies project, we use the Modern Chinese GSD treebank with 123k characters from Wikipedia.

The quantity of tagged Classical Chinese text is small compared to the much larger volume of unlabeled text. For unlabeled text, we use the Twenty-Four Histories corpus \citep{zinin-24hist}, containing about 40 million characters and written across 2000 years of Chinese history. Most of this text consists of biographies of prominent figures, written in Classical Chinese prose emulating the pre-Qin style. The dataset is released with a topic-modelling analysis of gender terms, which we do not use in this project.

One contested issue with Chinese text analysis is how to segment words. Neither Classical nor Modern Chinese separates words with spaces, so the definition of a word is indeterminate and different corpora disagree on segmentation conventions. In this work, we perform all analysis on the character level. Practically, this makes little difference, because the majority of the Kyoto corpus is monosyllabic, with the average word length being 1.1. This is not the only approach: \citet{nested-pos-tags} advocated for nested multi-level segmentation to preserve the internal structure of multi-character words.

\subsection{Part-of-speech tagging}

\begin{table*}[]
    \centering
    \begin{tabular}{|l|l|}
        \hline
        {\bf Tagger} & {\bf Accuracy} \\ \hline
        MeCab-Kanbun (Pre-Qin Chinese) & 0.61 \\ \hline
        UDPipe 1.2 (Pre-Qin Chinese) & 0.55 \\ \hline
        Stanford CoreNLP (Modern Chinese) & 0.57 \\ \hline
    \end{tabular}
    \caption{Character level part-of-speech tagging accuracy for taggers, evaluated on a small section of Twenty-Four-Histories text.}
    \label{tab:pos-results}
\end{table*}

We attempt to automatically process the Twenty-Four-Histories corpus with part-of-speech taggers. We use three different pretrained POS taggers: MeCab-Kanbun \citep{mecab-kanbun} and UDPipe 1.2 \citep{udpipe-1.2} which are trained on the Kyoto corpus of pre-Qin text, and Stanford CoreNLP \citep{stanford-corenlp} for Modern Chinese.

For evaluation, we pick an small sample of 112 characters from the Twenty-Four-Histories text and run it through each POS tagger. Because words in Classial Chinese are not delimited by spaces, tokenization is a necessary step prior to POS tagging, but we evaluate POS accuracy on a character-level, which allows us to evaluate the results even when the automatic tokenization is incorrect. The results are shown in Table \ref{tab:pos-results}.

Although all three taggers report over 90\% test accuracy on their respective test sets, we find that the accuracy is far worse (around 60\% accuracy) when applied to the Twenty-Four-Histories text. The two taggers trained on pre-Qin text make similar errors, but the Modern Chinese tagger makes very different errors; we next analyze the types of errors in more detail.

\subsubsection*{Pre-Qin POS tagger errors}

The most frequent type of error for MeCab-Kanbun and UDPipe was failing to recognize named entities and tagging them instead as either noun or verb. For example, the proper name 李白 (Li Bai, a poet) is tagged as 李 (plum, NOUN) + 白 (white, NOUN).

These errors are not unexpected because the Twenty-Four-Histories text, which are mostly biographies of prominent people, contains far more named entities than the pre-Qin training data, which is about Confucian philosophy. Furthermore, the people mentioned in the Twenty-Four-Histories did not exist during the pre-Qin period, explaining the taggers' difficulty in recognizing them.

POS tagging of the Twenty-Four-Histories text can likely be improved by annotating a portion of it for training data, or by combining a POS tagger with a database of named entities. We leave these ideas for future work.

\subsubsection*{Modern Chinese POS tagger errors}

\begin{table*}
    \centering
    \begin{tabular}{|l|l|l|}
    \hline
     & \textbf{Most common} & \textbf{Count} \\ \hline
    Flexible nouns & 子, 君, 天, 下, 礼, 王, 言, 道 & 128 \\ \hline
    Flexible verbs & 行, 知, 大, 事, 食, 使, 小, 至 & 140 \\ \hline
    Inflexible nouns & 人, 民, 夫, 国, 心, 日, 今, 士 & 135 \\ \hline
    Inflexible verbs & 曰, 有, 以, 为, 无, 如, 谓, 见 & 262 \\ \hline
    \end{tabular}
    \caption{Examples of flexible / inflexible nouns / verbs in Classical Chinese, and their counts.}
    \label{tab:flexible-examples}
\end{table*}

The Modern Chinese Stanford CoreNLP tagger, although also not satisfactory, produced very different errors from the ones trained on pre-Qin text. It correctly identified more of the proper names, but had a tendency to combine sequences of two characters as single disyllabic words. This makes sense because most words are disyllabic in Modern Chinese, but this is usually incorrect in Classical Chinese, a language where most words are monosyllabic. It also mistagged grammatical particles that are no longer in use, for example: 焉 (yan) and 矣 (yi).

Since none of the POS taggers showed adequate performance, we do not use their output for further experiments as originally planned. The Twenty-Four-Histories corpus is still valuable data for training unsupervised methods, such as word vectors.

\subsection{Aligned character embeddings}

We train character-level GloVe embeddings \citep{glove} on Classical and Modern Chinese. We use the full Twenty-Four Histories text (40M characters) for Classical Chinese, and the Chinese Wikipedia (1.2B characters) for Modern Chinese; 300-dimensional GloVe vectors are trained for both languages using the default settings.

For alignment, we follow \citet{hamilton-diachronic}, using the Orthogonal Procrustes method. Formally, we find an orthonormal rotation matrix $R$ that minimizes:
$$\min_R ||X_c - X_m R||_F^2,$$
where $X_c$ and $X_m$ are the GloVe vectors for Classical and Modern Chinese, respectively. The orthonormal requirement for $R$ ensures that the transformation preserves distances and dot products, which are essential for vector semantics. The solution to this optimization problem can be solved analytically using SVD \citep{svd-procrustes}.

\section{Experiments}

\begin{figure*}
    \centering
    \includegraphics[width=\linewidth]{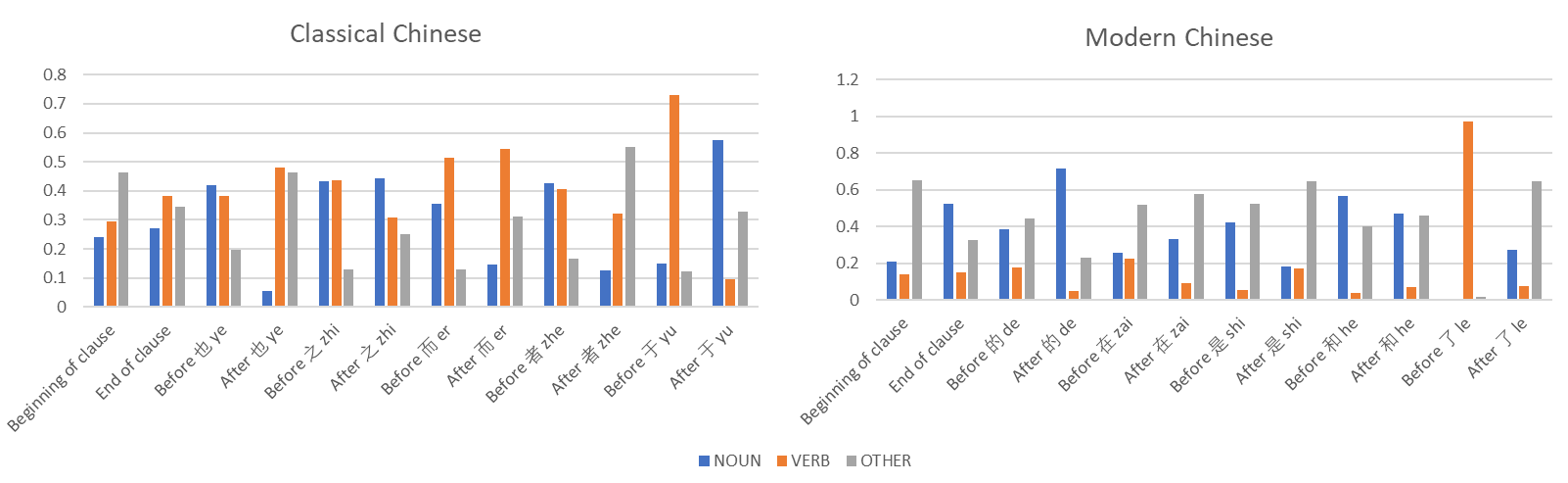}
    \caption{Distributions of POS tags noun, verb, and other, across 12 syntactic positions in Classical and Modern Chinese.}
    \label{fig:pos-dist-both}
\end{figure*}

\subsection{Flexible parts of speech}

\begin{table}[t]
    \centering
    \begin{tabular}{|l|l|}
    \hline
    \textbf{Position} & \textbf{Entropy} \\ \hline
    Beginning of clause & 1.060 \\ \hline
    End of clause & 1.088 \\ \hline
    Before 也 ye & 1.051 \\ \hline
    After 也 ye & 0.870 \\ \hline
    Before 之 zhi & 0.989 \\ \hline
    After 之 zhi & 1.069 \\ \hline
    Before 而 er & 0.974 \\ \hline
    After 而 er & 0.973 \\ \hline
    Before 者 zhe & 1.027 \\ \hline
    After 者 zhe & 0.955 \\ \hline
    Before 于 yu & 0.770 \\ \hline
    After 于 yu & 0.907 \\ \hline
    \textbf{Mean} & \textbf{0.978} \\ \hline
    \textbf{Median} & \textbf{0.982} \\ \hline
    \end{tabular}
    \caption{POS entropy of 12 syntactic positions in Classical Chinese.}
    \label{tab:pos-entropy-classical}
\end{table}

\begin{table}[t]
    \centering
    \begin{tabular}{|l|l|}
    \hline
    \textbf{Position} & \textbf{Entropy} \\ \hline
    Beginning of clause & 0.885 \\ \hline
    End of clause & 0.987 \\ \hline
    Before 的 de & 1.033 \\ \hline
    After 的 de & 0.727 \\ \hline
    Before 在 zai & 1.025 \\ \hline
    After 在 zai & 0.903 \\ \hline
    Before 是 shi & 0.856 \\ \hline
    After 是 shi & 0.894 \\ \hline
    Before 和 he & 0.811 \\ \hline
    After 和 he & 0.903 \\ \hline
    Before 了 le & 0.141 \\ \hline
    After 了 le & 0.836 \\ \hline
    \textbf{Mean} & \textbf{0.833} \\ \hline
    \textbf{Median} & \textbf{0.889} \\ \hline
    \end{tabular}
    \caption{POS entropy of 12 syntactic positions in Modern Chinese.}
    \label{tab:pos-entropy-modern}
\end{table}

We take all characters that appear at least 10 times in the Kyoto corpus, and consider a character as a noun if it appears more often as a noun than as a verb. We define a character to be flexible if it appears at least once as the other part of speech. Table \ref{tab:flexible-examples} gives examples of flexible nouns, flexible verbs, inflexible nouns, and inflexible verbs.

Of the 263 noun characters, 49\% of them are flexible, and of the 402 verb characters, 35\% of them are flexible. Although there is a high degree of flexibility, there are many examples of seemingly inflexible characters that appear hundreds of times as one word class and never as the other class.

This still does not falsify Bisang's claim that all words are flexible: using an inflexible noun in verb position may still be possible, just extremely rare so that in any given corpus the construction might be unattested. Usually, one would seek native speaker judgements of whether certain constructions are grammatical or not, but this is impossible because no human alive is a native speaker of Classical Chinese.

\subsection{Entropy of syntactic positions}

We next empirically measure how much the part of speech of a word is determined by syntactic position. There is no standard definition of syntactic position, so we first need to define a metric that can be applied to both Classical as well as Modern Chinese.

First, we extract the top five grammatical function words in both languages using the UD corpus, where the tags AUX, PART, CCONJ, SCONJ, AUX, and ADP are considered grammatical function words. In Classical Chinese, these are 也({\em ye}), 之({\em zhi}), 而({\em er}), 者({\em zhe}), 于({\em yu}); in Modern Chinese, these are 的({\em de}), 在({\em zai}), 是({\em shi}), 和({\em he}), 了({\em le}).

Next, we consider the word immediately before and after each grammatical function word to be a syntactic position (e.g. words immediately before 也 {\em ye}). This gives us 10 syntactic positions. We also take the first and last word of each clause, giving 12 syntactic positions in total (a clause here is any contiguous string of words without punctuation).

For each syntactic position, we count the POS of all words in that position in the corpus. All POS tags except for noun and verb are merged into a single ``other'' category. We now compute the POS entropy using the formula:
$$\textnormal{POS-entropy} = -\sum_{i=1}^3 p(POS_i) \log p(POS_i).$$
Lower entropy means that the part of speech is more predictable in the syntactic position.

The distributions of all syntactic positions are shown in Figure \ref{fig:pos-dist-both}, and the entropy values are in Tables \ref{tab:pos-entropy-classical} and \ref{tab:pos-entropy-modern}. There is evidence in Classical Chinese that the part of speech can be predicted somewhat by syntax, for example, the particle 于 ({\em yu}) usually comes after a verb and before a noun. However, the same can be said for Modern Chinese, and in fact the average entropy is actually lower in Modern Chinese than in Classical Chinese. Thus, Classical Chinese does not seem particularly unusual in the extent to which part of speech is determined by syntax.

\subsection{Noun ratio and concreteness}

According to \citet{bisang-precategorical}, most words in Classical Chinese can be used in either noun or verb positions, but not all words are attested in both positions because their semantics makes one position much more likely than the other. He proposes that abstract words are used more often as verbs, while concrete words are used more often as nouns. More specifically, he proposes a concreteness hierarchy: pronouns $>$ proper names $>$ human $>$ nonhuman $>$ abstracts.

We test this claim experimentally. There are no native speakers of Classical Chinese, so it is impossible to get native concreteness judgements for words. Instead, we use a dataset by \citet{yao-norms}, consisting of human ratings of 1,100 words in Modern Chinese on the following characteristics:

\begin{table}[]
    \centering
    \begin{tabular}{|l|l|l|}
        \hline
        {\bf Norm} & {\bf Pearson $\rho$} & {\bf p-value} \\ \hline
        Valence & -0.036 & 0.481 \\ \hline
        Arousal & -0.067 & 0.189 \\ \hline
        Concreteness & {\bf 0.138} & {\bf 0.006} \\ \hline
        Imageability & {\bf 0.127} & {\bf 0.013} \\ \hline
        Context availability & 0.106 & 0.038 \\ \hline
        Familiarity & 0.057 & 0.268 \\ \hline
    \end{tabular}
    \caption{Correlation of noun ratio with Modern Chinese norms from \citet{yao-norms}.}
    \label{tab:norms-correlation}
\end{table}

\begin{enumerate}
    \item Valence: degree to which the word feels positive vs negative.
    \item Arousal: degree to which the word feels calm vs exciting.
    \item Concreteness: how concrete or tangible is the object.
    \item Imageability: how easy it is to imagine a visual image of the object.
    \item Context availability: how easy it is to construct a sentence using the word.
    \item Familiarity: level of familiarity with the word.
\end{enumerate}

Words that are monosyllabic in Classical Chinese usually appear in multisyllablic compounds in Modern Chinese: for example, the word for ``friend'' is 友 ({\em you}) in Classical Chinese but is 朋友 ({\em pengyou}) in Modern Chinese. Thus we derive character-level ratings by taking the mean of the ratings of all Modern Chinese words containing the character.

For each character that appears at least 10 times in the Kyoto corpus, we compute the Pearson correlation between the noun ratio ($\frac{\#N}{\#N + \#V}$) and each rating dimension. We end up with 378 characters which have ratings. The correlations and p-values are shown in Table \ref{tab:norms-correlation}.

\begin{figure}
    \centering
    \includegraphics[width=0.8\linewidth]{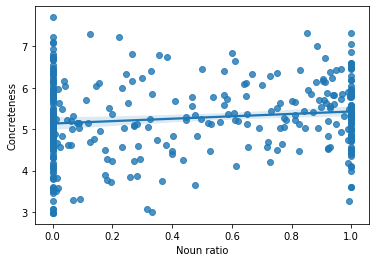}
    \caption{Relationship between noun ratio and concreteness norms (Pearson correlation $=0.138$, $p=0.006$).}
    \label{fig:concreteness}
\end{figure}

We find a statistically significant relationship between noun ratio and two of the rated characteristics: concreteness (Figure \ref{fig:concreteness}) and imageability. Although the relationship is weak, this supports the hypothesis that more concrete words occur more often as nouns. There is a weaker correlation with context availability, and no correlation with valence, arousal, or familiarity. Somewhat surprisingly, correlation is still detectable using Modern Chinese compounds, after being separated by over two thousand years of language change since the Classical pre-Qin period.

\subsection{Analysis of character semantic change}

\begin{table*}[t]
    \centering
    \begin{tabular}{|l|l|l|l|}
    \hline
    \textbf{Character} & \textbf{Meaning}                                                      & \textbf{Noun ratio} & \textbf{Semantic change} \\ \hline
    纳 na               & to receive / proper name                                              & 0                   & 1.427509                 \\ \hline
    术 shu              & art / skill / technology                                              & 1                   & 1.414176                 \\ \hline
    别 bie              & to separate / to leave / don't (MC)                               & 0                   & 1.396339                 \\ \hline
    适 shi              & to proceed / suitable                                                 & 0                   & 1.386691                 \\ \hline
    复 fu               & to return / repeat / copy                                             & 0                   & 1.38363                  \\ \hline
    与 yu               & particle (with)                                                       & 0                   & 1.383342                 \\ \hline
    异 yi               & different / other / unusual                                           & 0                   & 1.370065                 \\ \hline
    着 zhao             & to apply / to be affected / aspect particle (MC) & 0                   & 1.353356                 \\ \hline
    筑 zhu              & to build / ancient lute / proper name                                 & 0.166667            & 1.33977                  \\ \hline
    党 dang             & political party                                                       & 0.90625             & 1.333421                 \\ \hline
    \end{tabular}
    \caption{Top 10 characters with greatest semantic change from Classical to Modern Chinese. MC means a sense that is only used in Modern Chinese.}
    \label{tab:most-changed-chars}
\end{table*}

\begin{table*}[t]
    \centering
    \begin{tabular}{|l|l|l|l|}
    \hline
    {\bf Character} & {\bf Meaning}               & {\bf Noun ratio} & {\bf Semantic change} \\ \hline
    南 nan     & south                 & 0.96       & 0.721037        \\ \hline
    水 shui    & water                 & 1          & 0.739562        \\ \hline
    山 shan    & mountain              & 1          & 0.751939        \\ \hline
    病 bing    & sickness / to be sick & 0.483871   & 0.766609        \\ \hline
    食 shi     & food / to eat         & 0.299492   & 0.773277        \\ \hline
    河 he      & river                 & 1          & 0.778131        \\ \hline
    罪 zui     & crime                 & 0.781818   & 0.781502        \\ \hline
    死 si      & to die                & 0.252174   & 0.786616        \\ \hline
    城 cheng   & city                  & 1          & 0.790459        \\ \hline
    女 nu      & woman / female        & 0.935484   & 0.794239        \\ \hline
    \end{tabular}
    \caption{Top 10 characters with least semantic change from Classical to Modern Chinese.}
    \label{tab:least-changed-chars}
\end{table*}

Tables \ref{tab:most-changed-chars} and \ref{tab:least-changed-chars} show the ten characters with the most and least semantic change, measured by Euclidean distance.

Generally, meanings of characters broaden over time, such that senses in Classical Chinese are still in use in Modern Chinese, but there are additional novel senses that did not exist before. For example, the character 别 ({\em bie}) means ``to leave / to separate'' in Classical Chinese; this meaning is preserved in some compound words but is most commonly used as an imperative particle (``don't'') in Modern Chinese.

The characters with high semantic change are highly polysemous and are used in compounds in Modern Chinese. For example, 术 ({\em shu}) is rarely used by itself, but instead used in compounds like 技术 ({\em jishu}: technology / skill), 艺术 ({\em yishu}: art), 学术 ({\em xueshu}: learning / academic), and 手术 ({\em shoushu}: surgery). In Classical Chinese, this characters can take on multiple meanings, depending on context. In contrast, the characters with low semantic change have straightforward meanings and are valid standalone words in Modern Chinese.

\subsection{Factors that affect semantic change}

\begin{figure}
    \centering
    \includegraphics[width=0.8\linewidth]{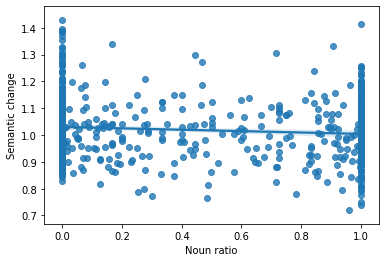}
    \caption{Semantic change as a function of noun ratio.}
    \label{fig:noun-ratio-semantic-change}
\end{figure}

\begin{figure}
    \centering
    \includegraphics[width=0.8\linewidth]{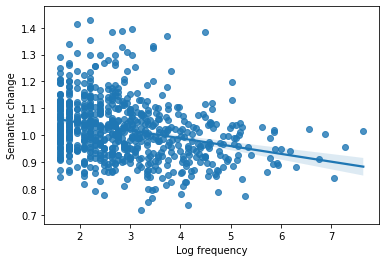}
    \caption{Semantic change as a function of log frequency.}
    \label{fig:frequency-semantic-change}
\end{figure}

\begin{table}[]
    \centering
    \begin{tabular}{|l|l|l|}
        \hline
        {\bf Metric} & {\bf Pearson $\rho$} & {\bf p-value} \\ \hline
        Noun ratio & -0.100 & 0.008 \\ \hline
        Log frequency & -0.282 & $<$0.001 \\ \hline
    \end{tabular}
    \caption{Correlation of semantic change with noun ratio and frequency.}
    \label{tab:semantic-change-correlation}
\end{table}

What are some factors that predict whether a word changes a lot or very little in meaning from Classical to Modern Chinese? In Table \ref{tab:semantic-change-correlation}, we examine two factors: noun ratio and frequency.

Noun ratio is negatively correlated with semantic change: in other words, verbs change more than nouns. The same effect has been found in English \citep{verb-sem-change}. Linguists disagree about the reasons for this effect; some proposals connect semantic change to the properties that verbs are more polysemous than nouns, or that verbs are acquired later than nouns.

Frequency is negatively correlated with semantic change; this effect has also been observed in English and is known as the law of conformity \citep{freq-sem-change}. However, the above paper pointed out that this effect is largely due to an artefact of word embeddings: infrequent words have noisier embeddings, which can be mistaken for greater change when semantic change is measured by distance in a vector space. Thus, it remains unclear whether Classical Chinese obeys the law of conformity.

\section{Conclusion}

Classical Chinese is a language with an extreme level of brevity and part-of-speech flexibility, making it unusual among the world's languages. In this paper, we use computational methods to conduct an investigation into various aspects of part-of-speech in Classical Chinese. We find that word classes are indeed quite flexible, with many words that can be used simultaneously in noun and verb positions. However, there are also many words that are attested only as a noun or as a verb.

We find evidence that concrete words tend to be used more in the noun position, and abstract words appear more in the verb position. Contrary to Bisang's claims, we do not find that the part-of-speech of a word can be inferred from its syntactic position. There is on average higher entropy in a given syntactic position in Classical Chinese than in Modern Chinese. Finally, we show that verbs undergo more semantic change than nouns, and more frequent words undergo less change, which agrees with previous work done in English.

Future work would benefit from improved part-of-speech tagging technology: this would unlock access to the vast amount of unlabelled Classical Chinese text for part-of-speech investigations, instead of only a small amount of tagged corpora. This data would better help us understand questions like whether a certain lexeme in verb position is in fact ungrammatical, or merely uncommon but still grammatical. Classical Chinese remains a valuable and under-utilized resource for studying historical semantic change -- it has over two millenia of text written in the same writing system, making it ideal for applying computational methods.

\section*{Acknowledgements}

The author thanks Sergei Zinin, Yang Xu, and Guillaume Thomas for their helpful guidance and suggestions.

\bibliography{acl2020}
\bibliographystyle{acl_natbib}

\end{CJK*}
\end{document}